\documentclass[10pt,twocolumn,letterpaper]{article}

\usepackage{cvpr}
\usepackage{times}
\usepackage{epsfig}
\usepackage{graphicx}
\usepackage{amsmath}
\usepackage{amssymb}
\usepackage{booktabs}
\usepackage{comment}
\usepackage{subfigure}
\usepackage[normalem]{ulem}
\usepackage{multirow}

\usepackage{listings}
\usepackage{color}

\usepackage{mathtools}

\definecolor{dkgreen}{rgb}{0,0.6,0}
\definecolor{gray}{rgb}{0.5,0.5,0.5}
\definecolor{mauve}{rgb}{0.58,0,0.82}

\usepackage{hyperref}
\hypersetup{colorlinks}

 \cvprfinalcopy % *** Uncomment this line for the final submission

\begin{document}

%%%%%%%%% TITLE
\title{Multi-Modal Face Anti-Spoofing Based on Central Difference Networks}

\author{Zitong Yu\textsuperscript{1}, Yunxiao Qin\textsuperscript{2}, Xiaobai Li\textsuperscript{1}, Zezheng Wang, Chenxu Zhao\textsuperscript{3},  Zhen Lei\textsuperscript{4}, Guoying Zhao\textsuperscript{1\thanks{denotes corresponding author}}\\
\normalsize{\textsuperscript{1}CMVS, University of Oulu  \qquad  \textsuperscript{2}Northwestern Polytechnical University}\\
\normalsize{\textsuperscript{3}Mininglamp Academy of Sciences, Mininglamp Technology \qquad
\textsuperscript{4}Authenmetric}\\
\tt\small \{zitong.yu, xiaobai.li, guoying.zhao\}@oulu.fi, \{qyxqyx\}@mail.nwpu.edu.cn\\
\tt\small\{zhaochenxu\}@mininglamp.com,\{zhen.lei\}@authenmetric.com
}

\maketitle

\begin{abstract}

Face anti-spoofing (FAS) plays a vital role in securing face recognition systems from presentation attacks. Existing multi-modal FAS methods rely on stacked vanilla convolutions, which is weak in describing detailed intrinsic information from modalities and easily being ineffective when the domain shifts (e.g., cross attack and cross ethnicity). In this paper, we extend the central difference convolutional networks (CDCN)~\cite{yu2020searching} to a multi-modal version, intending to capture intrinsic spoofing patterns among three modalities (RGB, depth and infrared). Meanwhile, we also give an elaborate study about single-modal based CDCN. Our approach won the first place in “Track Multi-Modal” as well as the second place in “Track Single-Modal (RGB)” of ChaLearn Face Anti-spoofing Attack Detection Challenge@CVPR2020~\cite{liu2020cross}. Our final submission obtains 1.02$\pm$0.59\% and 4.84$\pm$1.79\% ACER in “Track Multi-Modal” and “Track Single-Modal (RGB)”, respectively. The codes are available at  \href{https://github.com/ZitongYu/CDCN}{https://github.com/ZitongYu/CDCN}.

\end{abstract}

%%%%%%%%% BODY TEXT
\section{Introduction}

\thispagestyle{empty}

Face recognition has been widely used in many interactive artificial intelligence systems for its convenience (e.g., access control, face payment and device unlock). However, vulnerability to presentation attacks (PAs) curtails its reliable deployment. Merely presenting printed images or videos to the biometric sensor could fool face recognition systems. Typical examples of presentation attacks are print, video replay, and 3D masks. For the reliable use of face recognition systems, face anti-spoofing (FAS) methods are important to detect such presentation attacks.

\begin{figure}
\centering
\includegraphics[scale=0.43]{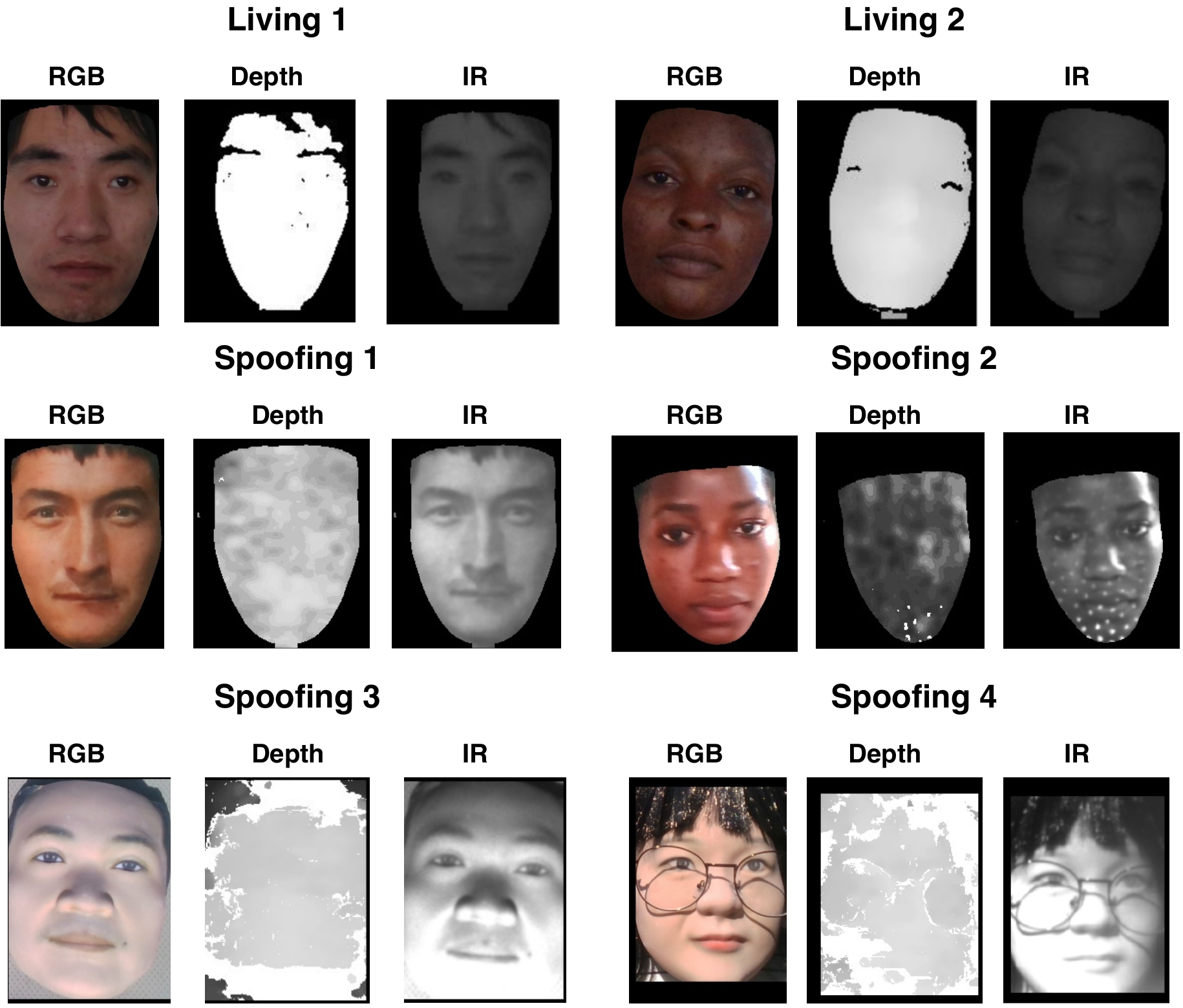}
  \caption{\small{
  Examples of living and spoofing faces from CASIA-SURF CeFA dataset~\cite{li2020casia}.}
  }
 
\label{fig:Figure1}
\end{figure}

%In this paper, we focus on developing effective methods for detecting planar attacks (e.g., print and video replay attacks). 

In recent years, several hand-crafted feature based~\cite{boulkenafet2015face,Boulkenafet2017Face,Pereira2012LBP,Komulainen2014Context,Peixoto2011Face,Patel2016Secure} and deep learning based~\cite{yu2020auto,wang2020deep,qin2019learning,Liu2018Learning,jourabloo2018face,yang2019face,Atoum2018Face,Gan20173D,george2019deep} methods have been proposed for presentation attack detection (PAD). On one hand, the classical hand-crafted descriptors (e.g., local binary pattern (LBP)~\cite{boulkenafet2015face}) leverage local relationship among the neighbours as the discriminative features, which is robust for describing the detailed invariant information (e.g., color texture, moir$\rm\acute{e}$ pattern and noise artifacts) between the living and spoofing faces. On the other hand, due to the stacked convolution operations with nonlinear activation, the convolutional neural networks (CNN) hold strong representation abilities to distinguish the bona fide from PAs. However, CNN based methods focus on the deeper semantic features, which are weak in describing detailed intrinsic information between living and spoofing faces and easily being ineffective when acquisition conditions varies (e.g., light illumination and camera type). In order to solve this issue, central difference convolutional networks (CDCN) is developed~\cite{yu2020searching} for single-modal (RGB) FAS task and achieves state-of-the-art performance on several benchmark datasets. Although the state-of-the-art single-modal FAS methods are robust in some existing testing protocols, it is still challenging when encountering new kinds of domain shift (e.g., cross ethnicity).

Recently, a large-scale cross-ethnicity face anti-spoofing dataset, the CASIA-SURF CeFA~\cite{li2020casia}, is established, which covers three ethnicities, three modalities, 1607 subjects, and 2D plus 3D attack types. Some typical examples are shown in Fig.~\ref{fig:Figure1}. The most challenging protocol 4 (simultaneously cross-attack and cross-ethnicity) is utilized for ChaLearn Face Anti-spoofing Attack Detection Challenge@CVPR2020~\cite{liu2020cross}. The baseline results in CASIA-SURF CeFA dataset~\cite{li2020casia} indicate: 1) multiple modalities (i.e., RGB, depth and infrared (IR)) fusion is more robust than using an arbitrary single modal, and 2) the multi-modal result, only 31.8$\pm$10.0\% ACER in protocol 4, is barely satisfactory. Hence it is necessary to explore more effective multi-modal FAS methods for cross-attack and cross-ethnicity testing.

Motivated by the discussions above, we first analyze how different modality influences the performance of CDCN. Then we extend CDCN to a multi-modal version, intending to capture intrinsic spoofing patterns among various modalities. Our contributions include:

 %Note that our work mainly focuses on detecting the planar attacks (e.g., print and replay).

\begin{itemize}
\setlength\itemsep{-0.1em}

    \item We are the first to utilize CDCN for depth and infrared modalities based FAS and analyze how CDCN performs with these two modalities. Besides considering CDCN as a single-modal network, we extend it to a multi-modal version, which captures rich discriminative clues among modalities and represents invariant intrinsic patterns across ethnicities and attacks.

    \item Our approach won the first place in “Track Multi-Modal”\footnote {\href{https://competitions.codalab.org/competitions/23318}{https://competitions.codalab.org/competitions/23318}} as well as the second place in “Track Single-Modal (RGB)”\footnote {\href{https://competitions.codalab.org/competitions/22151}{https://competitions.codalab.org/competitions/22151}} of ChaLearn Face Anti-spoofing Attack Detection Challenge@CVPR2020~\cite{liu2020cross}.
    
    %Comprehensive experiments are performed on six benchmark datasets to show our superior performance on both intra- and cross-dataset testing.
    
\end{itemize}

\section{Related Work}

In this section, we first introduce some recent progress in the single-modal FAS community; and then demonstrate few recent works about multi-modal FAS. Finally, classical convolution operators for vision tasks are presented.

\noindent\textbf{Single-Modal Face Anti-Spoofing.}\quad      
Traditional single-modal face anti-spoofing methods usually extract hand-crafted features from the RGB facial images to capture the spoofing patterns. Several classical local descriptors such as LBP~\cite{boulkenafet2015face,Pereira2012LBP}, 
SIFT~\cite{Patel2016Secure}, SURF~\cite{Boulkenafet2017Face_SURF}, HOG~\cite{Komulainen2014Context} and DoG~\cite{Peixoto2011Face} are utilized to extract frame level features while video level methods usually capture dynamic clues like dynamic texture~\cite{komulainen2012face}, micro-motion~\cite{siddiqui2016face} and eye blinking~\cite{Pan2007Eyeblink}. More recently, a few deep learning based methods
are proposed for both frame level and video level face anti-spoofing. For frame level methods \cite{yu2020searching,qin2019learning,Li2017An,Patel2016Cross,george2019deep,jourabloo2018face}, deep CNN models are utilized to extract features in a binary-classification setting. In contrast, auxiliary depth supervised FAS methods~\cite{Atoum2018Face,Liu2018Learning} are introduced to learn more detailed information effectively. On the other hand, several video level CNN methods are presented to exploit the dynamic spatio-temporal~\cite{wang2020deep,yang2019face,lin2018live} or rPPG~\cite{li2016generalized,Liu2018Learning,lin2019face,yu2019remote,yu2019remote2,shi2019atrial} features for PAD. Despite achieving state-of-the-art performance, single-modal methods are easily influenced by unseen domain shift (e.g., cross ethnicity and cross attack types) and not robust for challenging cases (e.g., harsh environment and realistic attacks).

\noindent\textbf{Multi-Modal Face Anti-Spoofing.}\quad  There are also few works for multi-modal face anti-spoofing. Zhang et al.~\cite{zhang2019dataset} take ResNet18 as the backbone and propose a three-stream network, where the input of each stream is RGB, Depth and IR face images, respectively. Then, these features are concatenated and passed to the last two residual blocks. Aleksandr et al.~\cite{parkin2019recognizing} also consider the similar fusion network with three streams. ResNet34 is chosen as the backbone and multi-scale features are fused at all residual blocks. Tao et al.~\cite{shen2019facebagnet} present a multi-stream CNN architecture called FaceBagNet. In order to enhance the local detailed representation ability, patch-level images are adopted as inputs. Moreover, modality feature erasing operation is designed to prevent overfitting and obtain more robust modal-fused features. All previous methods just consider standard backbone (ResNet) with stacked vanilla convolutions for multiple modalities, which might be weak in representing the intrinsic features between living and spoofing faces.

%\vspace{0.8em}
\noindent\textbf{Convolution Operators.}\quad  
The convolution operator is commonly used
in extracting basic visual features in deep learning framework. Recently extensions to the vanilla convolution operator have been proposed. In one direction, classical local descriptors (e.g., LBP \cite{ahonen2006face} and Gabor filters \cite{jain1991unsupervised}) are considered into convolution design. Representative works include Local Binary Convolution \cite{juefei2017local} and Gabor Convolution \cite{luan2018gabor}, which are proposed for saving computational cost and enhancing the resistance to the spatial changes, respectively. Recently, Yu et al. propose Central Difference Convolution (CDC) \cite{yu2020searching}, which is suitable for FAS task because of its excellent representation ability for detailed intrinsic patterns. Another direction is to modify the spatial scope for aggregation. Two related works are dialated convolution \cite{yu2015multi} and deformable convolution \cite{dai2017deformable}. However, these convolution operators are always designed for RGB modality, it is still unknown how they perform for depth and IR modalities.

In order to overcome the above-mentioned drawbacks and fill in the blank, we extend the state-of-the-art single-modal network CDCN to a multi-modal version for challenging cross-ethnicity and cross-attack FAS task.

\section{Methodology}
\label{sec:method}

In this section, we will first introduce CDC \cite{yu2020searching} as a preliminary in Section~\ref{sec:CDC}, then demonstrate our single-modal and multi-modal neural architectures in Section~\ref{sec:SCDCN} and Section~\ref{sec:MCDCN}, respectively. At last the supervision signals and loss functions are presented in Section~\ref{sec:fusion}.

\subsection{Preliminary: CDC}
\label{sec:CDC}

The feature maps and convolution can be represented in 3D shape (2D spatial domain and extra channel dimension) in modern deep learning frameworks. For simplicity, all convolutions in this paper are described in 2D while extension to 3D is straightforward.

\textbf{Vanilla Convolution.}\quad   
There are two main steps in the 2D spatial convolution: 1) \textsl{sampling} local receptive field region $\mathcal{R}$ over the input feature map $x$; 2) \textsl{aggregation} of sampled values via weighted summation. Hence, the output feature map $y$ can be formulated as

\begin{equation} 
y(p_0)=\sum_{p_n\in \mathcal{R}}w(p_n)\cdot x(p_0+p_n),
\label{eq:vanilla}
\end{equation}
where $p_0$ denotes current location on both input and output feature maps while $p_n$ enumerates the locations in $\mathcal{R}$. For instance, local receptive field region for convolution operation with 3$\times$3 kernel and dilation 1 is $\mathcal{R}=\left \{  (-1,-1),(-1,0),\cdots,(0,1),(1,1)  \right \}$.

\textbf{Central Difference Convolution.}\quad    For FAS task, the ‘discriminative’ and ‘robust’ features indicate fine-grained living/spoofing patterns and environment invariant clues, respectively. Local gradient operator (e.g., basic element in local binary pattern (LBP)~\cite{boulkenafet2015face}), as a residual and difference term, is able to capture rich detailed patterns and not easily affected by external changes.

Inspired by LBP~\cite{boulkenafet2015face}, we introduce central difference context into vanilla convolution to enhance its representation and generalization capacity. Similar to vanilla convolution, central difference convolution also consists of two steps, i.e., \textsl{sampling} and \textsl{aggregation}. The sampling step is similar to that in vanilla convolution while the aggregation step is different: central difference convolution prefers to aggregate the center-oriented gradient of sampled values. Thus Eq.~(\ref{eq:vanilla}) becomes

%It can be decomposed into two steps: perform depth-wise convolution to generate the difference map and then conduct the vanilla convolution.

\begin{equation} 
%y(p_0)=\sum_{p_n\in \mathcal{R}, p_n\neq (0,0)}w(p_n)\cdot (x(p_0+p_n)-x(p_0)),
y(p_0)=\sum_{p_n\in \mathcal{R}}w(p_n)\cdot (x(p_0+p_n)-x(p_0)).
\label{eq:central}
\end{equation}
When $p_n= (0,0)$, the gradient value always equals to zero with respect to the central location $p_0$ itself.

\begin{figure}
\centering
\includegraphics[scale=0.25]{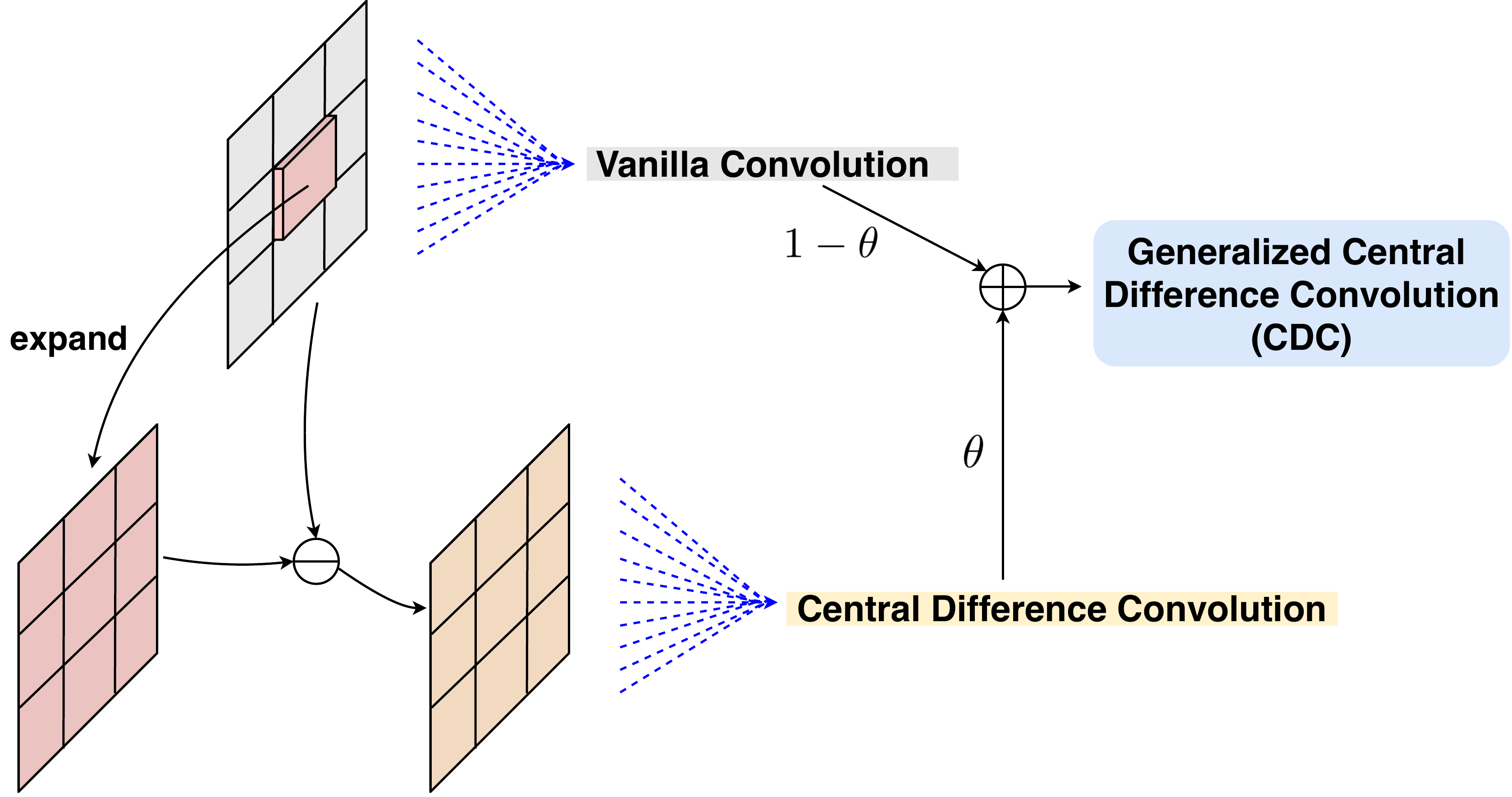}
  \caption{\small{
  Generalized central difference convolution (CDC). 
  }
  }
 
\label{fig:CDC}
\end{figure}

As both the intensity-level semantic information and gradient-level detailed message are crucial for distinguishing the living and spoofing faces, which indicates that combining vanilla convolution with central difference convolution might be a feasible manner to provide more robust modeling capacity. As illustrated in Fig.~\ref{fig:CDC}, we generalize central difference convolution as 

\begin{equation} 
\begin{split}
y(p_0)
=\theta \cdot \underbrace{\sum_{p_n\in \mathcal{R}}w(p_n)\cdot (x(p_0+p_n)-x(p_0))}_{\text{central difference convolution}}&\\
+ (1-\theta)\cdot \underbrace{\sum_{p_n\in \mathcal{R}}w(p_n)\cdot x(p_0+p_n)}_{\text{vanilla convolution}},& \\
\end{split}
\label{eq:CDC}
\end{equation}
where hyperparameter $\theta \in [0,1]$ tradeoffs the contribution between intensity-level and gradient-level information. The higher value of $\theta$ means the more importance of central difference gradient information. Similar to~\cite{yu2020searching}, we refer to this generalized central difference convolution as \textbf{CDC}.

\subsection{Single-Modal CDCN}
\label{sec:SCDCN}
\begin{figure*}
\centering
\includegraphics[scale=0.5]{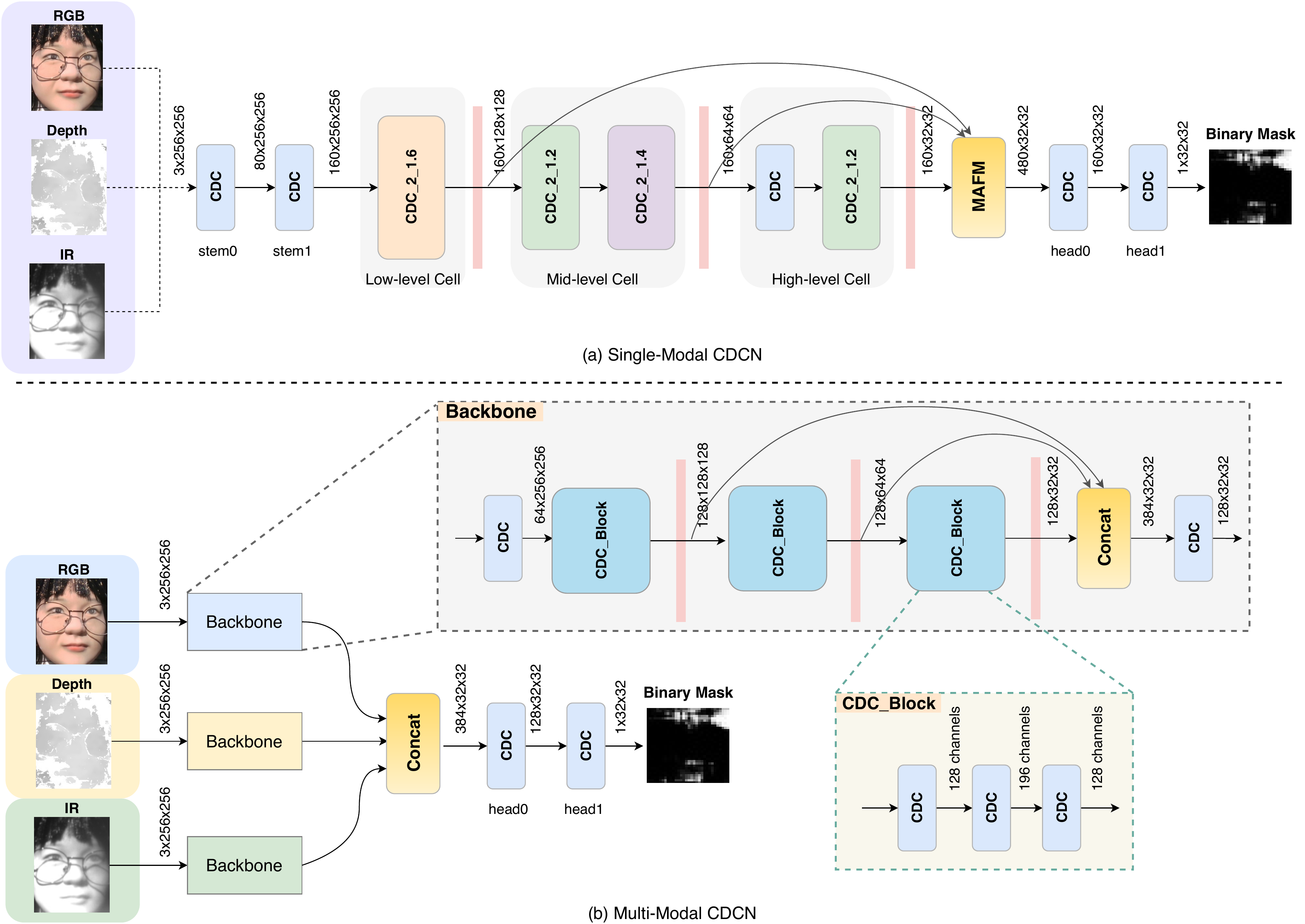}
  \caption{\small{
  The architecture of (a) single-model and (b) multi-modal CDCN. The red thin rectangle denotes a max pool layer with stride 2. `CDC\_2\_r' means using two stacked CDC to increase channel number with ratio r first and then decrease back to the original channel size. }
  }
 
\label{fig:networks}
\end{figure*}

We follow the similar configuration `CDCN++'~\cite{yu2020searching} as our single-modal backbone, including low-mid-high level cells and Multiscale Attention Fusion Module (MAFM). In the  consideration of the large-scale training data in CASIA-SURF CeFA dataset, we set the initial channel number as 80 instead of 64. The specific network is shown in Fig.~\ref{fig:networks}(a). Single-modal face image with size 256$\times$256$\times$3 is taken as the network input and the output is the predicted 32$\times$32 grayscale mask.

\subsection{Multi-Modal CDCN}
\label{sec:MCDCN}

We adopt the configuration `CDCN'~\cite{yu2020searching} as the backbone of each modality branch as we find the MAFM would drop the performance when using multi-modal fusion. As illustrated in Fig.~\ref{fig:networks}(b), the backbone network of each modality branch is not shared. Thus each branch is able to learn modality-aware features independently. The multi-level features from each modality branch are fused via concatenation. Finally, the two head layers aggregate the multi-modal features and predict the grayscale mask. 

As the feature-level fusion strategy might not be optimal for all protocols, we also try two other fusion strategies: 1) input-level fusion via concatenating three-modal inputs to 256$\times$256$\times$9 directly, and 2) score-level fusion via weighting the predicted score from each modality. For these two fusion strategies, the architecture of single-modal CDCN (see Fig.~\ref{fig:networks}(a)) is used. The corresponding ablation study will be shown in Section~\ref{sec:multitest}.

\subsection{Supervision}
\label{sec:fusion}

Compared with traditional guidance from the binary scalar score, pixel-wise supervision~\cite{george2019deep} helps to learn more discriminative patterns between living and spoofing faces. As a result, our network prefets to predict 32$\times$32 grayscale mask instead of traditional scalar score. In terms of ground truth label, we generate the binary mask via simply set the non-zero pixel value to `1' because the intensity values of non-face background have already been `0' in CASIA-SURF CeFA dataset.

\begin{figure}
\centering
\includegraphics[scale=0.48]{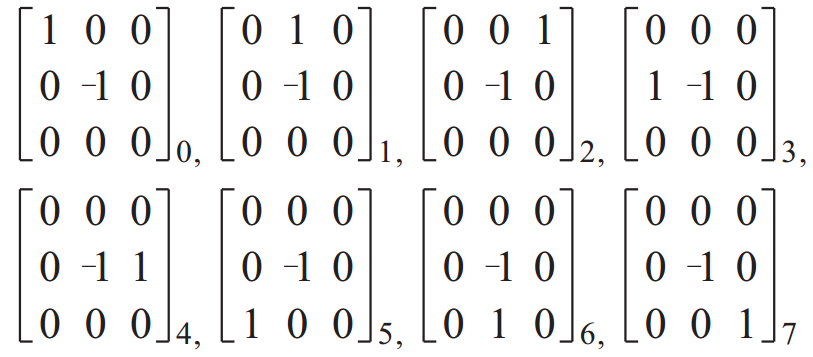}
  \caption{\small{
  The kernel $K_{n}^{CDL}$ in contrastive depth loss.. 
  }
  }
 
\label{fig:CDL}
\end{figure}

\newcommand{\tabincell}[2]{\begin{tabular}{@{}#1@{}}#2\end{tabular}}
\begin{table*}
\centering
\caption{Albation study of the hyperparameter $\theta$ with RGB modality.}

\scalebox{0.75}{\begin{tabular}{|c|c|c|c|c|c|c|c|c|c|c|}
\hline
\multirow{2}{*}{Single-Modal CDCN} &\multicolumn{3}{c|}{Protocol 4@1} &\multicolumn{3}{c|}{Protocol 4@2}&\multicolumn{3}{c|}{Protocol 4@3} &{Overall} \\
\cline{2-11} &\tabincell{c}{APCER(\%)} &\tabincell{c}{BPCER(\%)} &\tabincell{c}{ACER(\%)} &\tabincell{c}{APCER(\%)}&\tabincell{c}{BPCER(\%)}&\tabincell{c}{ACER(\%)}&\tabincell{c}{APCER(\%)}&\tabincell{c}{BPCER(\%)}&\tabincell{c}{ACER(\%)} & \tabincell{c}{ACER(\%)}\\
\hline
$\theta$=0.5
& 12.61 & 4.0 & 8.31 & 6.67 & 2.0 & \textbf{4.33} & 4.56 & 8.5 & 6.53 & 6.39 \\
\hline
$\theta$=0.6
& 11.67 & 8.0 & 9.83 & 10.56 & 3.0 & 6.78 & 3.89 & 5.0
 & 4.44 & 7.02 \\
\hline
$\theta$=0.7
& 12.83 & 1.25 & 7.04 & 13.33 & 2.0 & 7.67 & 3.72 & 3.0 & \textbf{3.36} & \textbf{6.02}\\
\hline
$\theta$=0.8
& 14.33 & 1.5 & 7.92 & 10.0 & 6.25 & 8.13 & 3.83 & 7.25 & 5.54 & 7.19 \\
\hline
$\theta$=0.9
& 11.17 & 2.5 & \textbf{6.83} & 21.33 & 5.75 & 13.54 & 3.56 & 7.5 & 5.53 & 8.63 \\
\hline
\end{tabular}
}
\label{tab:theta2}

\end{table*}

\begin{table*}
\centering
\caption{Results of Single-Modal CDCN ($\theta$=0.7) with different modalities.}

\scalebox{0.75}{\begin{tabular}{|c|c|c|c|c|c|c|c|c|c|c|}
\hline
\multirow{2}{*}{Modality} &\multicolumn{3}{c|}{Protocol 4@1} &\multicolumn{3}{c|}{Protocol 4@2}&\multicolumn{3}{c|}{Protocol 4@3} &{Overall} \\
\cline{2-11} &\tabincell{c}{APCER(\%)} &\tabincell{c}{BPCER(\%)} &\tabincell{c}{ACER(\%)} &\tabincell{c}{APCER(\%)}&\tabincell{c}{BPCER(\%)}&\tabincell{c}{ACER(\%)}&\tabincell{c}{APCER(\%)}&\tabincell{c}{BPCER(\%)}&\tabincell{c}{ACER(\%)} & \tabincell{c}{ACER(\%)}\\
\hline
RGB
&12.83 & 1.25 & 7.04 & 13.33 & 2.0 & 7.67 & 3.72 & 3.0 & 3.36 & 6.02\\
\hline
Depth
& 5.22 & 1.25 & 3.24 & 2.72 & 0.5 & \textbf{1.61} & 4.94 & 1.75 & \textbf{3.35} & \textbf{2.73} \\
\hline
IR
& 1.56 & 1.0 & \textbf{1.28} & 27.72 & 0.25 & 13.99 & 29.56 & 0.5 & 15.03 & 10.1\\
\hline

\end{tabular}
}
\label{tab:modality}

\end{table*}

\begin{table*}[!htbp]
\centering
\caption{Best submission result in Track Single-Modal (RGB).}

\scalebox{0.75}{\begin{tabular}{|c|c|c|c|c|c|c|c|c|c|c|}
\hline
\multirow{2}{*}{Method} &\multicolumn{3}{c|}{Protocol 4@1} &\multicolumn{3}{c|}{Protocol 4@2}&\multicolumn{3}{c|}{Protocol 4@3} &{Overall} \\
\cline{2-11} &\tabincell{c}{APCER(\%)} &\tabincell{c}{BPCER(\%)} &\tabincell{c}{ACER(\%)} &\tabincell{c}{APCER(\%)}&\tabincell{c}{BPCER(\%)}&\tabincell{c}{ACER(\%)}&\tabincell{c}{APCER(\%)}&\tabincell{c}{BPCER(\%)}&\tabincell{c}{ACER(\%)} & \tabincell{c}{ACER(\%)}\\
\hline
SD-Net ~\cite{li2020casia}
&- & - & - & - & - & - & - & - & - & 35.2$\pm$5.8\\
\hline
\textbf{Ours (Single-Modal)}
& 11.17 & 2.5 & \textbf{6.83} & 6.67 & 2.0 & \textbf{4.33} & 3.72 & 3.0 & \textbf{3.36} & \textbf{4.84$\pm$1.79} \\
\hline

\end{tabular}
}
\label{tab:singleSOTA}

\end{table*}

For the loss function, mean square error loss $\mathcal{L}_{MSE}$ is utilized for pixel-wise supervision, which is formulated:
\begin{equation} 
\mathcal{L}_{MSE}=\frac{1}{H\times W}\sum_{i\in H,j\in W}(B_{pre(i,j)}-B_{gt(i,j)})^2,
\label{eq:MSE}
\end{equation}
where $H,W$ denote the height and width of the binary mask, respectively, and $B_{pre}$ and $B_{gt}$ mean the predicted grayscale mask and ground truth binary mask, respectively. Moreover, for the sake of fine-grained supervision needs in FAS task, contrastive depth loss (CDL) $\mathcal{L}_{CDL}$~\cite{wang2020deep} is considered to help the networks learn more detailed features. CDL can be formulated as 
\begin{equation}\small 
\mathcal{L}_{CDL}=\frac{\sum_{i\in H,j\in W,n\in N}(K_{n}^{CDL}\odot B_{pre(i,j)}-K_{n}^{CDL}\odot B_{gt(i,j)})^2}{H\times W\times N},
\label{eq:CDL}
\end{equation}
where $K_{n}^{CDL}$ is the $n$-th contrastive convolution kernel, and $N$ denotes the kernel numbers. The details of the kernels ($N=8$) can be found in Fig.~\ref{fig:CDL}. Finally, the overall loss $L_{overall}$ can be formulated as $\mathcal{L}_{overall}=\mathcal{L}_{MSE}+\mathcal{L}_{CDL}$.

\section{Experiments}

\label{sec:experiemnts}
In this section, extensive experiments are performed to demonstrate the effectiveness
of our method. In the following, we sequentially
describe the employed datasets \& metrics (Sec. \ref{sec:dataset}), implementation details (Sec. \ref{sec:Details}), results (Sec. \ref{sec:singletest} - \ref{sec:multitest}) and visualization (Sec. \ref{sec:Analysis}).

\subsection{Datasets and Metrics}

\label{sec:dataset}
\textbf{CASIA-SURF CeFA Dataset~\cite{li2020casia}.} 
CASIA-SURF CeFA aims to provide with the largest up to date face anti-spoofing dataset to allow for the evaluation of the generalization performance cross-ethnicity and cross-attacks. It consists of 2D and 3D attack subsets. For the 2D attack subset, it includes print and video-reply attacks, and three ethnicites (African, East Asian and Central Asian) with two attacks (print face from cloth and video-replay). Each ethnicity has 500 subjects. Each subject has one real sample, two fake samples of print attack captured in indoor and outdoor, and 1 fake sample of video-replay. In total, there are 18000 videos (6000 per modality). 

There are four evaluation protocols in CASIA-SURF CeFA for cross-ethnicity, cross-attack, cross-modality, and cross-ethnicity \& cross-attack testing. In this paper, our experiments are all conducted on the most challenging protocol 4 (cross-ethnicity \& cross-attack), which has been utilized for ChaLearn Face Anti-spoofing Attack Detection Challenge@CVPR2020.

\textbf{Performance Metrics.}\quad
Three metrics, i.e., Attack Presentation Classification Error Rate (APCER), Bona Fide Presentation Classification Error Rate (BPCER), and Average Classification Error Rate (ACER)~\cite{ACER} are utilized for performance comparison. They can be formulated as
\begin{equation}
\begin{split}
& APCER  = \frac{FP }{TN + FP}, \\
& BPCER  = \frac{FN }{FN + TP}, \\
& ACER  = \frac{APCER +BPCER}{2}, \\
\label{eq:ACER}
\end{split}
\end{equation}
where $FP$, $FN$, $TN$ and $TP$ denote the false positive, false negative, true negative and true positive sample numbers, respectively. ACER is used to determine the final ranking in ChaLearn Face Anti-spoofing Attack Detection Challenge@CVPR2020.

\subsection{Implementation Details}

\label{sec:Details}

Our proposed method is implemented with Pytorch. In the training stage, models are trained with Adam optimizer and the initial learning rate and weight decay are 1e-4 and 5e-5, respectively. We train models with 50 epochs while learning rate halves every 20 epochs. The batch size is 8 on a P100 GPU. In the testing stage, we calculate the mean value of the predicted grayscle map as the final score.

\begin{table*}[!htbp]
\centering
\caption{Ablation study of fusion strategies for multi-modal CDCN. We only report the results tried in the FAS challenge.}

\scalebox{0.75}{\begin{tabular}{|c|c|c|c|c|c|c|c|c|c|}
\hline
\multirow{2}{*}{Modality} &\multicolumn{3}{c|}{Protocol 4@1} &\multicolumn{3}{c|}{Protocol 4@2}&\multicolumn{3}{c|}{Protocol 4@3}\\
\cline{2-10} &\tabincell{c}{APCER(\%)} &\tabincell{c}{BPCER(\%)} &\tabincell{c}{ACER(\%)} &\tabincell{c}{APCER(\%)}&\tabincell{c}{BPCER(\%)}&\tabincell{c}{ACER(\%)}&\tabincell{c}{APCER(\%)}&\tabincell{c}{BPCER(\%)}&\tabincell{c}{ACER(\%)} \\
\hline
Feature-level fusion
&0.33 & 0.5 & \textbf{0.42} & 5.89 & 3.25 & 4.57 & 4.22 & 3.25 & 3.74 \\
\hline
Input-level fusion
& 0.5 & 3.75 & 2.13 & 5.67 & 1.5 & 3.58 & 2.61 & 3.25 & 2.93 \\
\hline
Score-level fusion
& - & - & - & 1.39 & 0.75 & \textbf{1.07} & 1.44 & 1.75 & \textbf{1.6} \\
\hline

\end{tabular}
}
\label{tab:multifusion}

\end{table*}

\begin{table*}[!htbp]
\centering
\caption{Best submission result in Track Multi-Modal.}

\scalebox{0.75}{\begin{tabular}{|c|c|c|c|c|c|c|c|c|c|c|}
\hline
\multirow{2}{*}{Method} &\multicolumn{3}{c|}{Protocol 4@1} &\multicolumn{3}{c|}{Protocol 4@2}&\multicolumn{3}{c|}{Protocol 4@3} &{Overall} \\
\cline{2-11} &\tabincell{c}{APCER(\%)} &\tabincell{c}{BPCER(\%)} &\tabincell{c}{ACER(\%)} &\tabincell{c}{APCER(\%)}&\tabincell{c}{BPCER(\%)}&\tabincell{c}{ACER(\%)}&\tabincell{c}{APCER(\%)}&\tabincell{c}{BPCER(\%)}&\tabincell{c}{ACER(\%)} & \tabincell{c}{ACER(\%)}\\
\hline
PSMM-Net~\cite{li2020casia}
&33.3 & 15.8 & 24.5 & 78.2 & 8.3 & 43.2 & 50.0 & 5.5 & 27.7 & 31.8$\pm$10.0\\
\hline
\textbf{Ours (Multi-Modal)}
& 0.33 & 0.5 & \textbf{0.42} & 1.39 & 0.75 & \textbf{1.07} & 1.44 & 1.75 & \textbf{1.6} & \textbf{1.02$\pm$0.59} \\
\hline

\end{tabular}
}
\label{tab:multiSOTA}

\end{table*}

\subsection{Single-Modal Testing}

\label{sec:singletest}

In this subsection, we give the ablation study about the hyperparameter $\theta$ with RGB modality firstly. Then based on the optimal $\theta$ for CDCN, we test depth and IR modalities. Finally, we summarize our best submission results in “Track Single Modal (RGB)” on ChaLearn Face Anti-spoofing Attack Detection Challenge@CVPR2020.

\textbf{Impact of $\theta$ with RGB modality.}\quad 
As shown in Table~\ref{tab:theta2}, the best overall performance (ACER=6.02\%) is achieved when $\theta=0.7$, which is consistent with the evidence in ~\cite{yu2020searching}. As for the sub-protocols, $\theta=0.9$, $\theta=0.5$ and $\theta=0.7$ obtain the lowest ACER in protocol 4@1 (6.83\%), 4@2 (4.33\%) and 4@3 (3.36\%), respectively.

\begin{figure*}[!htbp]
\centering
\includegraphics[scale=0.27]{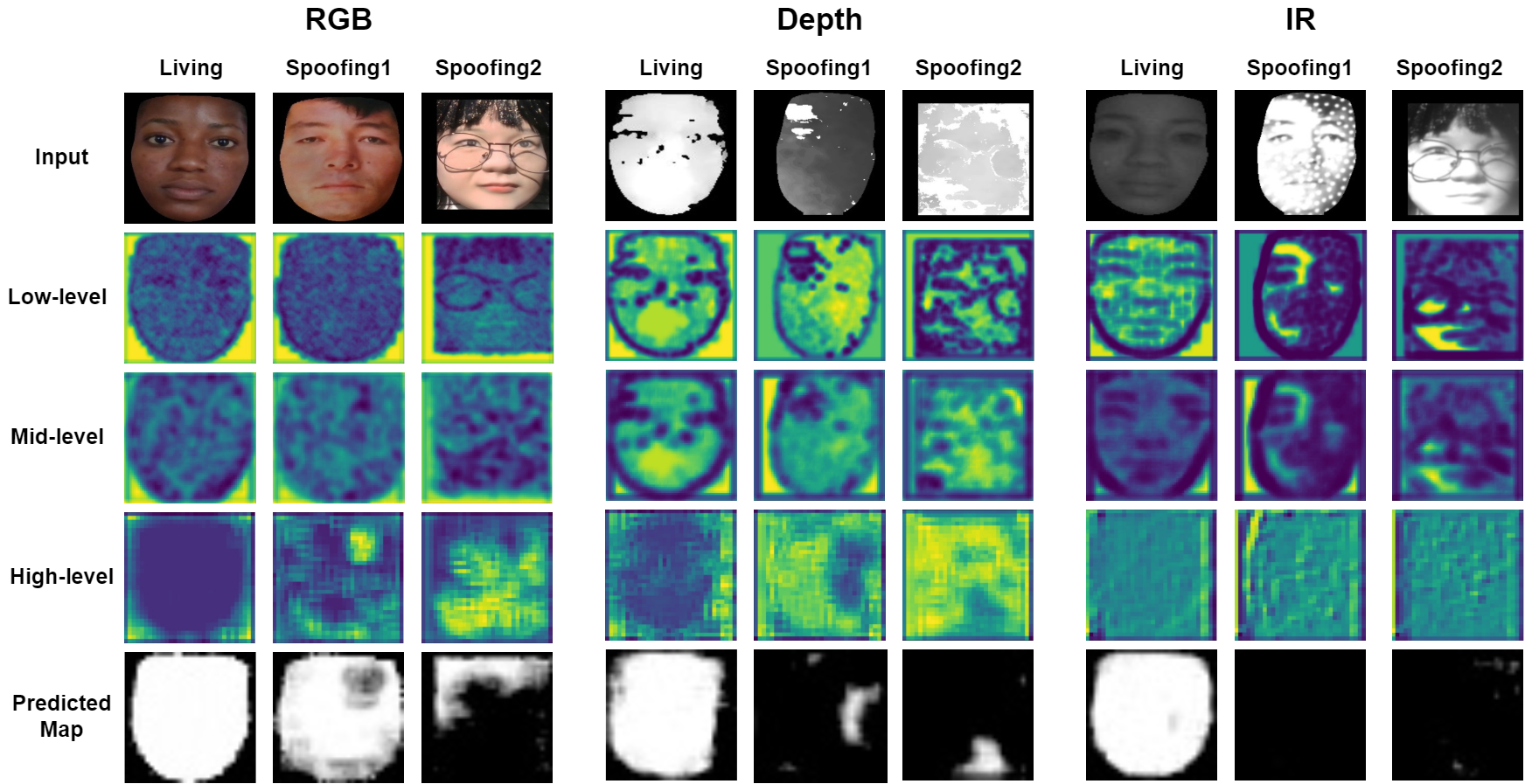}

  \caption{\small{
  Visualization of CDCN with three modalities. }
  }
 
\label{fig:visual}
\end{figure*}

\textbf{Results of Depth and IR modalities.}\quad 
Table~\ref{tab:modality} shows the results of different modalities using single-modal CDCN when $\theta=0.7$.  It is surprising that the performance varies a lot across modalities. The IR modality performs the best in protocol 4@1 (testing without Africa) but the worst in protocol 4@2 and 4@3 (testing with Africa), indicating that the IR modality generalizes poorly for unseen Africa ethnicity. Compared with RGB and IR modalities, the depth modality is more robust and discriminative in most cases (e.g., print attacks in testing stage) because the 3D depth shape is quite distinguishable between living and print faces. The excellent overall performance indicates central difference convolution is not only suitable for RGB modality, but also for IR and depth modalities.

\textbf{Best Submission Result in Track Single-Modal (RGB).}\quad 
Our best submission result (4.84$\pm$1.79\% ACER) is shown in Table~\ref{tab:singleSOTA}, which wins the second place in Track Single-Modal (RGB) on ChaLearn Face Anti-spoofing Attack Detection Challenge@CVPR2020. This final result is combined with the best sub-protocols results (i.e., $\theta=$0.9, 0.5 and 0.7, respectively).

\subsection{Multi-Modal Testing}

\label{sec:multitest}

In this subsection, three fusion strategies are studied in multi-modal testing. Then the best submission results in Track Multi-Modal will be presented.

\textbf{Multi-Modal Fusion Strategies.}\quad 
As shown in Table~\ref{tab:multifusion}, our proposed multi-modal CDCN (i.e., feature-level fusion with three modalities) achieves the lowest ACER (0.42\%) in protocol 4@1. When using the concatenated inputs with three modalities (input-level fusion), the CDCN could obtain comparable performance with the single-modal results in Table~\ref{tab:modality}. However, it still causes the performance drops compared with the best single-modal results (i.e., IR modality for protocol 4@1, depth modality for protocol 4@2 and protocol 4@3). It also reflects the issue for both feature- and input-level fusion, i.e., simple fusion with concatenation might be sub-optimal because it is weak in representing and selecting the importance of modalities. It is worth exploring more effective fusion methods (e.g., attention mechanism for modalities) in future. 

Based on the prior results in Table~\ref{tab:modality}, we weight the results of RGB and depth modalities averagely as the score-level fusion (i.e., $fusion\_score=0.5*RGB\_score+0.5*depth\_score$). As shown in Table~\ref{tab:multifusion} (the third row), this simple ensemble strategy helps to boost the performance significantly. Compared with single-depth modality, score-level fusion gives 0.54\% and 1.13\% ACER improvements for protocol 4@2 and 4@3, respectively.

\textbf{Best Submission Result in Track Multi-Modal.}\quad 
Table~\ref{tab:singleSOTA} shows our best submission result (1.02$\pm$0.59\% ACER), which wins the first place in “Track Multi-Modal” on ChaLearn FAS Attack Detection Challenge@CVPR2020. This final result is combined with the best sub-protocols results (i.e., feature-level fusion for protocol 4@1 while score-level fusion for protocol 4@2 and 4@3).

\subsection{Feature Visualization}
 \label{sec:Analysis}

The visualizations of CDCN with three modalities are shown in Fig.~\ref{fig:visual}. On one hand, it is clear that the low-level, mid-level and high-level features in CDCN are distinguishable between living and spoofing faces among all three modalities. In terms of low-level features, the living have more detailed texture (especially in IR modality). As for the high-level features, the living face regions are purer and plainer while the spoofing ones are with more spoofing/noise patterns. 

On the other hand, depth and IR modalities are complementary to RGB modality and helpful for robust liveness detection. We can see from the last row in Fig.~\ref{fig:visual} that CDCN fails to detect spoofing1 only using RGB input while spoofing1 could be accurately detected by depth or IR inputs.

\section{Conclusion}
\label{sec:conc}

In this paper, we give an elaborate study about the applications of central difference convolutional networks (CDCN)~\cite{yu2020searching} for multiple modalities in face anti-spoofing (FAS) task. The experimental results indicate the effectiveness of CDCN for both single-modal and multi-modal FAS. The proposed approach wins the first place in “Track Multi-Modal” as well as the second place in “Track Single-Modal (RGB)” of ChaLearn Face Anti-spoofing Attack Detection Challenge@CVPR2020.

\section{Acknowledgement}

This work was supported by the Academy of Finland for project MiGA (Grant 316765), ICT 2023 project (Grant 328115), Infotech Oulu and the Chinese National Natural Science Foundation Projects (Grant No. 61876178). As well, the authors acknowledge CSC–IT Center for Science, Finland, for computational resources.

{\small
\bibliographystyle{ieee_fullname}
\bibliography{egbib}
}

\end{document}